\documentclass[10pt,twocolumn,letterpaper]{article}

\usepackage{cvpr}              

\usepackage{graphicx}
\usepackage{amsmath}
\usepackage{amssymb}
\usepackage{booktabs}
\usepackage{multirow}
\usepackage{makecell}
\usepackage{siunitx}   
\usepackage{amstext}   
\usepackage{tabularx}

\usepackage[pagebackref,breaklinks,colorlinks]{hyperref}

\usepackage[capitalize]{cleveref}
\crefname{section}{Sec.}{Secs.}
\Crefname{section}{Section}{Sections}
\Crefname{table}{Table}{Tables}
\crefname{table}{Tab.}{Tabs.}

\def\cvprPaperID{*****} 
\def\confName{CVPR}
\def\confYear{2025}

\begin{document}

\title{Interactive Multimodal Fusion with Temporal Modeling}
\author{
Jun Yu$^1$, Yongqi Wang$^1$, Lei Wang$^1$\thanks{Corresponding author}, Yang Zheng$^1$, Shengfan Xu$^2$\\
$^1$University of Science and Technology of China\\
$^2$Macau University of Science and Technology\\
\tt\small \{harryjun, wangl\}@ustc.edu.cn\\
\tt\small \{wangyongqi,zhengyang\}@mail.ustc.edu.cn \\
\tt\small \{eyki1010\}@163.com \\
}

\maketitle

\begin{abstract}
This paper presents our method for the estimation of valence-arousal (VA) in the 8th Affective Behavior Analysis in-the-Wild (ABAW) competition. Our approach integrates visual and audio information through a multimodal framework. The visual branch uses a pre-trained ResNet model to extract spatial features from facial images. The audio branches employ pre-trained VGG models to extract VGGish and LogMel features from speech signals. These features undergo temporal modeling using Temporal Convolutional Networks (TCNs). We then apply cross-modal attention mechanisms, where visual features interact with audio features through query-key-value attention structures. Finally, the features are concatenated and passed through a regression layer to predict valence and arousal. Our method achieves competitive performance on the Aff-Wild2 dataset, demonstrating effective multimodal fusion for VA estimation in-the-wild.
\end{abstract}

\section{Introduction}
In human-computer interaction and psychological research, accurate emotion recognition is crucial for effective communication\cite{usman2018using, KHARE2024102019}\cite{kollias2021analysing}\cite{kollias2020analysing}. Valence-arousal (VA) estimation\cite{kollias2022abaw}, which captures the nuanced nature of emotions, has garnered significant attention\cite{kollias2024distribution}. Unlike categorical emotion classification\cite{kollias2023multi}, VA estimation predicts two dimensions: valence (positivity/negativity) and arousal (excitement/calmness), offering a more comprehensive understanding of emotional states\cite{Sandbach2012StaticAD}\cite{Kollias2025}\cite{kollias20247th}.
Recent advances in deep learning and computer vision have greatly enhanced VA estimation from facial images and videos. Sophisticated neural architectures like CNNs and Transformers can extract high-level features of facial expressions. Moreover, integrating multimodal data\cite{Zafeiriou2017AffWildVA, Kollias_2019}\cite{kollias2019expression}\cite{kollias2019deep}, including audio and text, has further improved the accuracy and robustness of VA estimation systems by leveraging complementary information from different modalities.
However, VA estimation in real-world settings remains challenging due to factors like varying lighting, head poses, occlusions, and diverse facial expressions. The subjective nature of emotions and cultural differences in expression add to this complexity\cite{kollias2019face}. To address these challenges, we propose a multimodal data fusion method that integrates visual and audio information.
Our approach uses pre-trained models (VGGish\cite{hershey2017cnn} and IResnet\cite{duta2020improved}\cite{Deng_2022}) to extract features from both modalities, processes them with multi-scale temporal convolutional networks (TCNs)\cite{bai2018empirical}, and fuses them through cross-modal attention mechanisms. The main contributions are:

1. A novel multimodal fusion approach using pre-trained audio and video models to extract dynamic features.

2. Multi-scale TCNs to capture temporal dependencies while maintaining a large receptive field and preserving temporal resolution

3. Cross-modal attention mechanisms to effectively capture relationships between audio and visual features, enhancing the model's ability to recognize emotional states.

4. Robust performance in real-world scenarios, with significant improvements in VA estimation accuracy compared to baseline approaches.

\section{Related Work}
\textbf{Valence-Arousal Estimation:}
Valence-arousal (VA) estimation has been a focal point in emotion recognition research, aiming to capture the continuous dimensions of emotional states\cite{kolliasadvancements}\cite{kollias20246th}\cite{kollias2021affect}\cite{zafeiriou2017aff}. Traditional approaches relied on handcrafted features\cite{meng2022valence}\cite{zhang2021continuous} from facial images or audio signals, which were then fed into machine learning models. However, these methods were limited by the complexity and subjectivity of feature extraction.
The advent of deep learning marked a significant turning point, with convolutional neural networks (CNNs)\cite{poria2016convolutional,zhang2020m} becoming the backbone of VA estimation systems. These networks automatically learn hierarchical features from raw data, enabling more accurate and robust emotion recognition. The Aff-Wild2 dataset, annotated with valence and arousal labels, has been pivotal in training and evaluating CNN-based models. Researchers have explored various CNN architectures, such as VGG\cite{hershey2017cnn}, ResNet\cite{duta2020improved}, and DenseNet\cite{iandola2014densenet}, to extract spatial features from facial images. These spatial features are further enhanced by incorporating temporal information through techniques like Long Short-Term Memory (LSTM) networks\cite{graves2012long} or 3D CNNs\cite{maturana2015voxnet}, which capture the dynamic changes in facial expressions over time.

\textbf{Multimodal Fusion:}
The integration of multimodal data has proven highly effective in boosting VA estimation system performance. Audio-visual approaches combine features from facial images and speech signals, leveraging the complementary information from each modality\cite{kollias2023abaw}. For instance, the PDEM\cite{wagner2023dawn} fine-tuned on Wav2Vec\cite{baevski2020wav2vec} pre-trained audio models predicts arousal, valence, and dominance from speech. Combined with CNN-extracted visual features, these audio features offer a more comprehensive emotional state representation.
Recent research has explored Transformers\cite{vaswani2017attention} for multimodal fusion in VA estimation. Their self-attention mechanisms make Transformers adept at capturing long-range dependencies and inter-modal interactions. Some studies concatenate features from different modalities and apply Transformer layers for fusion\cite{wang2023tetfn}. Cross-attention mechanisms also allow each modality to focus on relevant features in others, further improving fusion effectiveness\cite{li2024multi}.

\textbf{Temporal Modeling:}
Temporal dynamics are crucial in VA estimation as emotions evolve over time. Various techniques have been developed to model these temporal dependencies.
TCNs are popular for capturing temporal patterns through dilated convolutions, allowing a large receptive field with minimal computational cost. They've been successfully applied to encode temporal information in both visual and audio modalities\cite{yu2024multimodal}.
RNNs\cite{medsker2001recurrent}, especially LSTMs\cite{graves2012long} and GRUs\cite{chung2014empirical}, are also widely used for temporal modeling. These networks maintain a hidden state to capture information from previous time steps. However, compared to TCNs and Transformers, RNNs face vanishing gradient issues and have limited parallelization capabilities.

\textbf{Recent Advances:}
Recent progress in self-supervised learning has introduced new possibilities for enhancing VA estimation. Notably, Masked Autoencoders (MAEs)\cite{he2022masked} have shown promise in learning robust visual representations. By pre-training on large-scale datasets\cite{kollias2021distribution} and fine-tuning for specific tasks, MAEs reduce reliance on labeled data and improve model generalization.
Moreover, cascaded cross-attention mechanisms within Transformer architectures have been proposed to iteratively refine multimodal features\cite{praveen2021cross}\cite{duan2021audio}. These methods apply cross-attention multiple times, enabling each modality to progressively focus on the most relevant features from other modalities. This recursive fusion strategy has demonstrated improved performance in capturing the intricate relationships between different modalities.
In summary, while significant progress has been made in VA estimation through deep learning, multimodal fusion, and advanced temporal modeling techniques, there remains room for improvement. Our work contributes to this ongoing research by proposing a novel integration of these elements, specifically tailored to address the challenges of VA estimation in-the-wild.
\begin{figure*}[ht]
\centering
\includegraphics[width=\linewidth]{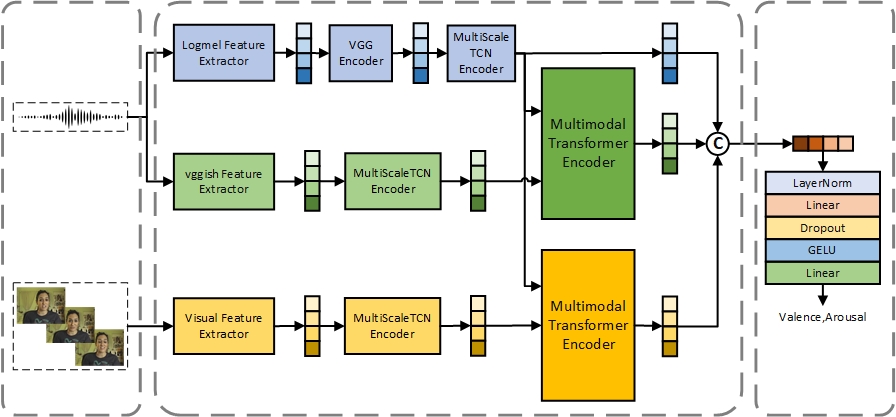}
\caption{Our proposed framework for VA estimation.}
\label{Figure 1}
\end{figure*}
\section{Proposed approach}
\subsection{Overview}
Our proposed method for Valence-Arousal (VA) estimation in the 8th Affective Behavior Analysis in-the-wild (ABAW) competition employs a multimodal framework with three key components: a visual branch, two audio branches, and two cross-modal attention fusion modules. The visual branch uses a pre-trained ResNet model to extract spatial features from facial images. The audio branches leverage pre-trained VGG models to extract VGGish and LogMel features from speech signals. Each modality's features are independently processed by multi-scale Temporal Convolutional Networks (TCNs) to capture temporal dynamics at various scales. We then apply cross-modal attention mechanisms, where visual features serve as queries to interact with audio features (keys and values), creating joint representations that capture inter-modal relationships. Finally, the concatenated features from all modalities are passed through a regression layer to predict valence and arousal. This architecture effectively integrates visual and audio information to enhance the accuracy and robustness of VA estimation in unconstrained environments.

\subsection{Visual Branch}
The visual branch employs a pre-trained ResNet model to extract spatial features $F_v$  from facial images. 
\begin{equation}
F_v = \text{ResNet}(I)
\end{equation}
where $I$denotes the input facial image. The ResNet model is selected for its robustness in capturing detailed facial features pertinent to emotion recognition. Each video frame is processed to generate spatial features that capture the visual information essential for understanding facial expressions.

\subsection{Audio Branches}
Audio Branch 1: VGGish Model
The first audio branch employs the VGGish model, a pre-trained audio representation model fine-tuned on the Audioset dataset. It processes raw audio waveforms to extract features $F_{vgg}$ that capture relevant acoustic features for emotion recognition.
\begin{equation}
F_{vgg} = \text{VGGish}(A)
\end{equation}
where $A$ represents the input audio signal.
The extracted features are then processed by a TCN to model temporal dynamics in the audio signals.

Audio Branch 2: LogMel Processing
The second audio branch processes audio signals by first computing LogMel spectrograms $S$:
\begin{equation}
S = \text{LogMel}(A)
\end{equation}
which provide a frequency-based representation of the audio. Features $F_{lm}$ are then extracted:
\begin{equation}
F_{lm} = \text{VGG}(S)
\end{equation}

\subsection{Temporal Modeling with Multi-Scale TCNs}
After extracting spatial features from both visual and audio modalities, we employ multi-scale Temporal Convolutional Networks (TCNs) to capture temporal dynamics within each modality. Specifically, we utilize TCNs with kernel sizes of 3, 5, and 7 to process the features, then concatenate the resulting features to form a comprehensive temporal representation.
For the visual branch:
\begin{equation}
F_v^{tcn} = \text{Concat}(\text{TCN}_3(F_v), \text{TCN}_5(F_v), \text{TCN}_7(F_v))
\end{equation}
For the audio branches:
\begin{equation}
F_{vgg}^{tcn} = \text{Concat}(\text{TCN}3(F{vgg}), \text{TCN}5(F{vgg}), \text{TCN}7(F{vgg}))
\end{equation}
\begin{equation}
F_{lm}^{tcn} = \text{Concat}(\text{TCN}3(F{lm}), \text{TCN}5(F{lm}), \text{TCN}7(F{lm}))
\end{equation}
Multi-scale TCNs are chosen for their ability to capture temporal dependencies at various scales through dilated convolutions. This approach allows the network to maintain a large receptive field while preserving temporal resolution, effectively modeling both short-term and long-term temporal patterns in the data.

\subsection{Cross-Modal Attention Fusion}
Visual and audio features are fused through cross - modal attention mechanisms, with visual features as queries and audio features as keys and values:
\begin{equation}
\text{Attention}(Q, K, V) = \text{Softmax}\left(\frac{QK^T}{\sqrt{d_k}}\right)V
\end{equation}
where $Q$ is the query (visual features), $K$ and $V$ are the key and value (audio features), and $d_k$is the dimension of the key vectors.
\begin{equation}
F_{v\_{vgg}} = \text{Attention}(F_v^{tcn}, F_{vgg}^{tcn}, F_{vgg}^{tcn})
\end{equation}
\begin{equation}
F_{v\_lm} = \text{Attention}(F_v^{tcn}, F_{lm}^{tcn}, F_{lm}^{tcn})
\end{equation}
This enables visual features to focus on relevant audio features, creating a joint representation that captures the relationships between different modalities.

\subsection{Regression Layer}
The final component of our architecture is the regression layer, which maps the concatenated feature vector $F_{concat}$ to the continuous dimensions of valence and arousal.The features from all modalities are concatenated as follows:
\begin{equation}
F_{concat} = [F_v^{tcn}; F_{v\_vgg}; F_{v\_lm}]
\end{equation}
This layer comprises several sub-layers designed to transform the high-dimensional feature representation into final predictions while enhancing model expressiveness and generalization through regularization and non-linear activation.

First, the concatenated features undergo layer normalization to stabilize the training process and improve convergence:
\begin{equation}
F_{norm} = \text{LayerNorm}(F_{concat})
\end{equation}
Next, the normalized features pass through a linear layer followed by a GELU activation function, introducing non-linearity while mitigating vanishing gradient issues:
\begin{equation}
F_{hidden} = \text{GELU}(W_1 \cdot F_{norm} + b_1)
\end{equation}
A dropout layer is then applied to prevent overfitting by randomly deactivating a fraction of the elements in $F_{hidden}$ to zero during training:
\begin{equation}
F_{dropout} = \text{Dropout}(F_{hidden})
\end{equation}
Finally, another linear transformation is applied to obtain the predicted valence and arousal values:
\begin{equation}
\hat{y} = W_2 \cdot F_{dropout} + b_2
\end{equation}

\subsection{Training Strategy}
Our model undergoes a multi - stage training process. Initially, the visual and audio branches are pre - trained on large - scale datasets to acquire robust feature representations. The ResNet50 - IR model is pre - trained on MS - Celeb - 1M and then fine - tuned on FER+, while the VGGish model is pre - trained on Audioset. Subsequently, the TCNs in each branch are trained to capture temporal dynamics specific to the ABAW task. Finally, the entire architecture, including the Transformer encoder and regression layer, is trained end - to - end on the Aff - Wild2 dataset. This comprehensive strategy ensures each model component learns the necessary features and interactions for accurate VA estimation in - the - wild.

\begin{table}[htbp]
\centering
\caption{The dimensions of features.}
\label{tab:feature_dimensions}
\begin{tabular}{ccc}
\toprule
\textbf{Feature Modality} & \textbf{Dimension} & \textbf{Description} \\ \midrule
VGGish & 128 & A \\
LogMel & 128 & A \\
IResNet-50 & 512 & V \\ \bottomrule
\end{tabular}
\end{table}

\section{Experiment}
In this section, we provided an overview of the dataset, evaluation protocol, and experimental results used in the study.
\subsection{Dataset and Evaluation}
\textbf{Affwild2 dataset.} The Aff-Wild2 dataset is a cornerstone for affective behavior analysis in-the-wild, particularly for Valence-Arousal (VA) estimation. As an extension of the original Aff-Wild dataset, Aff-Wild2 doubles the number of video frames and subjects, enhancing the diversity of behaviors and individuals represented. This dataset provides comprehensive annotations for three main behavioral tasks: expression recognition, action unit detection, and VA estimation. For VA estimation, it offers frame-based continuous annotations of valence and arousal, capturing nuanced emotional states in natural settings.
Aff-Wild2 consists of 564 videos with approximately 2.8 million frames, featuring 554 subjects with diverse demographics, including various ages, ethnicities, and nationalities. The dataset encompasses a wide range of environmental conditions and situational variations, making it suitable for developing and evaluating models for robust emotion recognition in real-world scenarios. The inclusion of both audio and video data enables multimodal approaches to VA estimation, allowing for a more comprehensive analysis of emotional expressions.
The frame-based annotations in Aff-Wild2 are valuable for VA estimation, as they provide detailed, per-frame measurements of valence and arousal. This granularity supports the development of models that capture the dynamic and subtle changes in emotional states over time, which is essential for accurate VA estimation in-the-wild.

\textbf{Evaluation.}To estimate the VA, we computed the Concordance Correlation Coefficient (CCC) separately for arousal and valence. 
\begin{equation}
P = 0.5 \times (CCC_{\text{arousal}} + CCC_{\text{valence}})
\end{equation}
where $CCC_\text{arousal}$ is the $CCC$ for arousal and $CCC_\text{valence}$ 
is the $CCC$ for valence.
\subsection{Results on Validation set}
Table \ref{tab:results} presents the performance comparison of our proposed method and the baseline on the Aff-Wild2 dataset across six validation folds. The results demonstrate that our method consistently outperforms the baseline for both valence and arousal estimation tasks. For valence estimation, our method achieves higher CCC scores across all folds compared to the baseline. Similarly, for arousal estimation, our method shows superior performance with higher CCC scores in all folds. These results highlight the effectiveness of our proposed approach in capturing the complex emotional dimensions from multimodal data in-the-wild.


\begin{table}[htbp]
\centering
\caption{Valence and Arousal Scores for Different Validation Sets}
\label{tab:results}
\setlength{\tabcolsep}{12pt} 
\begin{tabular}{@{}>{\raggedright}p{3.5cm}c@{\hspace{1.2em}}c@{}}
\toprule
\textbf{Validation Set} & \textbf{Valence} & \textbf{Arousal} \\
\midrule
Baseline & 0.24 & 0.20 \\
Fold-0 & 0.3672 & 0.6633 \\
Fold-1 & 0.4793 & 0.6159 \\
Fold-2 & 0.5968 & 0.6682 \\
Fold-3 & 0.5692 & 0.6228 \\
Fold-4 & 0.6045 & 0.6528 \\
Fold-5 & 0.4046 & 0.6184 \\
\bottomrule
\end{tabular}
\end{table}

\section{Conclusion}
This paper presents a comprehensive and robust approach to valence-arousal (VA) estimation for the 8th Affective Behavior Analysis in-the-Wild (ABAW) competition. Our method integrates visual and audio information through a multimodal framework, leveraging pre-trained models to extract features from both modalities and employing multi-scale temporal convolutional networks (TCNs) to capture temporal dynamics. Cross-modal attention mechanisms are utilized to effectively fuse these features, creating joint representations that capture the relationships between different modalities. The concatenated features are then passed through a regression layer to predict valence and arousal.
Our approach demonstrates significant improvements in VA estimation accuracy compared to baseline methods, achieving competitive performance on the Aff-Wild2 dataset. This comprehensive method not only integrates visual and audio information effectively but also addresses the challenges of VA estimation in unconstrained environments. The results highlight the effectiveness of our proposed approach in capturing the complex emotional dimensions from multimodal data in-the-wild.
Future work could explore further refinement of the multimodal fusion techniques, incorporation of additional data sources, and enhancement of the temporal modeling components to continue advancing the state-of-the-art in this critical area of affective computing.

{\small
\bibliographystyle{ieee_fullname}
\bibliography{egbib}
}

\end{document}